# Digital Twin Based Disaster Management System Proposal: DT-DMS

**Özgür Doğan**
Department of Computer Enginnering
Muğla Sıtkı Koçman University
Muğla, Turkey
ozgurdogan3@posta.mu.edu.tr

**Oğuzhan Şahin**
Department of Computer Enginnering
Muğla Sıtkı Koçman University
Muğla, Turkey
oguzhansahin3@posta.mu.edu.tr

**Enis Karaarslan**
Department of Computer Enginnering
Muğla Sıtkı Koçman University
Muğla, Turkey
enis.karaarslan@mu.edu.tr

*Abstract*— The damage and the impact of natural disasters are becoming more destructive with the increase of urbanization. Today's metropolitan cities are not sufficiently prepared for the pre and post-disaster situations. Digital Twin technology can provide a solution. A virtual copy of the physical city could be created by collecting data from sensors of the Internet of Things (IoT) devices and stored on the cloud infrastructure. This virtual copy is kept current and up to date with the continuous flow of the data coming from the sensors. We propose a disaster management system utilising machine learning called DT-DMS is used to support decision-making mechanisms. This study aims to show how to educate and prepare emergency centre staff by simulating potential disaster situations on the virtual copy. The event of a disaster will be simulated allowing emergency centre staff to make decisions and depicting the potential outcomes of these decisions. A rescue operation after an earthquake is simulated. Test results are promising and simulation scope is planned to be extended.

## I. INTRODUCTION

Metropolitan cities are becoming more crowded and their transportation, telecommunication, electricity networks, and natural gas networks are becoming more complex day by day. Natural disasters are a reality and these big cities have to be ready for pre and post-disaster situations. Turkey is no exception.

The number of natural disasters such as earthquakes and their destructiveness are expected to increase in the future. The most recent earthquake in Elazığ showed that disaster management practices and the tools to train staff for pre and post-disaster situation are not sufficient. The fundamental elements of post-disaster interventions were late and communication infrastructures collapsed.

Digital twin technology can be used as a solution. Digital twin can be defined as the bridge between the physical and virtual world. The data received from physical assets are transferred to the digital twin. Digital twin technology can be used to build smarter buildings and infrastructures like transportation, telecommunication, electricity, and natural gas networks. Such a solution should be developed before any crisis to allow emergency staff to be trained to meet the surmountable challenges of a natural disasters, especially an earthquake.

In the next section, fundamentals of digital twin solution are explained. Related works are provided in the third section. The proposal of the system is explained in the fourth section followed by the implementation in section five. The results and conclusions are included in final section.

## II. FUNDAMENTALS

### A. Disaster Management

Today's technology is not yet sufficient in predicting when disasters will take place [1]. However, today's technology is sufficient to supply means which enables us to take measures before and after the disaster to reduce catastrophic effects of any incident on human lives and property. Disaster management can be used for the following:

- Prevent and reduce the impact of the disasters,
- Respond to the events in a timely, fast and effective manner.
- Create a new and safer environment for the victims.

Disaster management can be examined as pre and post disasters. In pre-disaster, the following actions can be taken to reduce risk and harm.

- Determining regions that exist on the fault line,
- Informing and raising awareness of the society about disaster hazard and risk,
- Improving disaster management capacity and capability,
- Installation of IoT devices in the buildings and in the critical infrastructures such as transportation, telecommunication, electricity and natural gas networks.

In a post-disaster situation, timely response by allocating resources to the most affected areas is one of the most critical aspects to reduce human suffering and number of casualties. In this context, gathering current information becomes the most crucial thing. Information/data about the affected areas can be gathered by using the IoT devices that are installed during the pre-disaster period. This data can be transferred to the rescue teams and could be used to allocate scarce resources to the most affected and areas to facilitate efficient relief operations. Drones, satellites, and similar devices could also be used to collect information on conditions of the disaster region. The data that is collected from these devices could be used in making decisions about the establishment of the following:

- number and nature of rescue teams,
- kind of equipment,
- mode of transportation,
- location of the relief camps.

There is a need for a decision support system that combines all the collected data and helps in producing an insight into the disaster.

### B. Smart City and Smart Nation

The migration of people to metropolitan cities for certain reasons makes management difficult in those regions. The population in cities in Europe was 50 percent in 1950, now it is over 75 percent [2]. The urban population will be approximately 6.7 billion in 2050 [3]. As a result of this urbanization, systems generate big data. The smart city [4] concept is about collecting these data and using them in decision support systems. Technologies such as IoT, AI (artificial intelligence), cloud computing [5], and blockchain [6] can be used. These all can be used to form a smart nation as shown in Fig. 1.

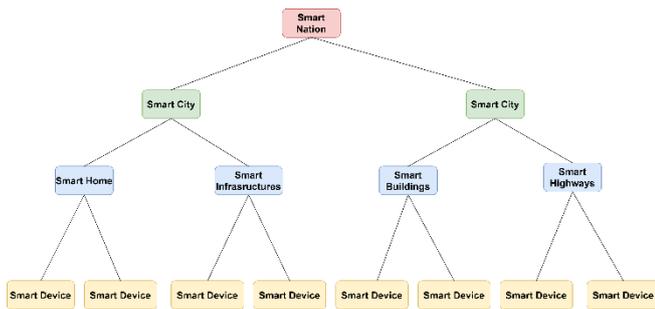

**Fig.1:** Smart Nation Concept

### C. Digital Twin

Digital twin technology is based on creating a virtual copy of an object, process, or human being in a virtual world. This virtual copy is kept current with continuous incoming data.

The concept of digital twin is not new. This technology was used for the first time in the 1960s by NASA (National Aeronautics and Space Administration). NASA has created virtual duplicate systems for a space mission to test its equipment and estimate potential costs. The "digital twin" term was introduced in 2002 by NASA's advisor Dr. Michael Grieves. It was also developed in the US Air Force and applied to the aircraft body models in 2013. Digital twin technology has become one of the most promising technology trends with the widespread usage of IoT (Internet of things).

Essential simulations and examinations to test a system, mechanism or a tool can be done easily with this technology reducing potential costs and manpower as well as avoiding harms to human life. The digital twin consists of three main concepts; physical assets, sensors, and virtual copy. The virtual copy displays the data from the IoT devices as 2D/3D or non-visual. Decision-making process of humans or machines can become easier using this display.

As open-source software has become more popular in the computer industry [7], it is also sensible to use a variety of these open-source protocols, frameworks, and tools for digital twins and IoT. Eclipse Ditto, AllJoyn, and IoTivity frameworks can be used. Communication protocols that can be used are; MQTT (Message Queue Telemetry), AMQP (Advanced Message Queueing Protocol), COAP (Constrained Application Protocol), XMPP (Extensible Messaging and Present Protocol) [8] and RabbitMQ.

### D. Digital Twin Communication

One of the most important features of humankind is the ability to communicate with each other. Life proceeds in a certain order with the help of this communication. Roads, buildings and infrastructures in the smart city which were created exist individually in the system. They can act by each other's feedback.

The connectivity mostly occurs between IoT devices. The number of IoT connected devices increased by 285% since 2015 and the number of these devices is around 38.5 billion [9]. The sectors such as manufacturing, logistics, health, agriculture, and automotive use IIoT (Industrial Internet of Things) often. Disruptions, delays, and stoppages of jobs can cause substantial losses. Therefore, communication is an important component for the IIoT. In smart cities, the communication must not be interrupted between digital twins for the sustainability of management and life.

The fourth period of the industry has begun (Industry 4.0) which is based on CPS (cyber-physical systems). CPS is a hybrid system that has a physical and virtual environment in communication with each other. In CPS everything is interconnected wirelessly. Smart factories, autonomous cars, quadcopters, etc. are substantial examples of CPS.

These devices are in constant communication with each other and also connected to the internet. There are lots of different ways to make these connections such as RFID, ZigBee, WPAN, WSN, DSL, UMTS, GPRS, WiFi, WiMax, LAN, WAN, 3G, etc [10]. These types are given in Table 1.

**Table 1.** IoT Connectivity Types

| Type | Radio Frequency | Range | Data Transfer Speed |
|---|---|---|---|
| Bluetooth | 2.4 GHz-2.5 GHz | 0,5 m-100 m | $\leq 24 Mbps$ |
| Wi-Fi | 2.4 GHz-5 GHz | 45 m-91 m | $\leq 250 Mbps$ |
| Zigbee | 860 Mhz-2.4 GHz | 10 m-100 m | $\leq 250 Kbps$ |
| GPRS | Diff. in each reg. | 8 km-40 km | 56-114 Kbps |
| 3G | Diff. in each reg. | 8 km-40 km | 0,8-2 Mbps |
| 4G | Diff. in each reg. | 8 km-40 km | 0.2-1 Gbps |
| 5G | Diff. in each reg. | 8 km-40 km | $\geq 1 Gbps$ |

### III. RELATED WORKS

There are mainly touristic implementations in the smart city concept. Newcastle, Dubai, Qatar and Singapore started to form their digital twins. Qatar and Singapore have the aim to use digital twins in a touristic sense. In the three-dimensional virtual copy of the city, it also aimed for people to go on the streets and see touristic places [11].

However, using digital twins for disaster management is rare. Digital twin technology is used for so many different projects and studies in the literature. [12-15]. For example, it is used for controlling floods in Newcastle by monitoring the amount of water on the roads [16].

However, there is a lack of academic studies that can be used for better solutions and standardization. Only a recent study [17] addressed this idea and proposed a "disaster city" concept. Our study is an attempt to develop a proof of concept how to implement Digital Twin on a disaster management setting where further studies can be built upon. This prototype will address a variety of disaster (earthquake, storm, fire etc.) types and collect data from a variety of sources such as electricity, natural gas and communication infrastructures of the buildings are planned to be controlled.

## IV. SYSTEM PROPOSAL

The proposed system aims to develop essential measures for pre- and post-disaster environment by performing various simulations on a virtual copy. These measures include information on as buildings, transportation, telecommunication, natural gas and electricity networks. This system could also be used real-time to collect data following a disaster to decide how rescue operations are managed by understanding of the occupancy rate of critical facilities such as hospitals.

Digital twin architecture is shown in Fig. 2. Sensors collect data from the physical world and sent it to the virtual environment which is stored in the cloud. Continuous flow of data is the most essential and crucial requirement of the system. This data can be collected from many sources such as IoT sensors, social media, data resources from government and municipality. This data will be stored in database systems on the cloud.

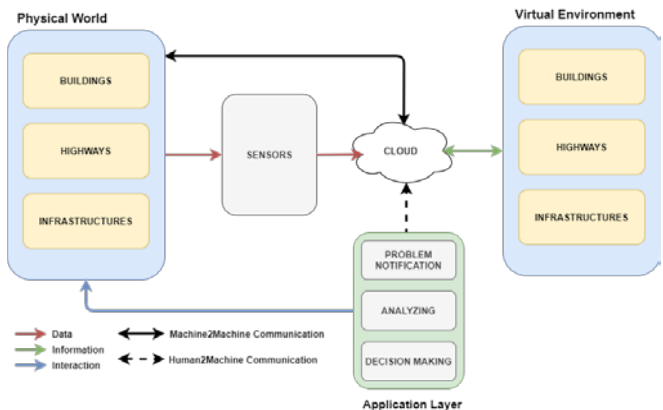

**Fig. 2:** Digital Twin Architecture

There are two main actors of the system that are the staff at the disaster management centers and the rescue teams. Organizations can use this system from the rescue centers and the teams can use these data during the intervention. Also in the post-disaster scenario; buildings and rescue center information of the relevant streets can be checked by clicking on the icon of buildings and the rescue center. In the infrastructures section, electricity, water, and telecommunication facilities can be inspected. A timeline button is used in order to analyze the effects of the disaster on time. The system can be run in two modes:

- Education mode: Based on the decision of the management staff, the simulation is made and possible outcomes are reported.
- Estimating mode: Decision support system runs in several different choices and tries to find the best decision for the best outcome.

The action diagram of the system is in Fig 3. Diagram consists of three components; server, IoT Devices, and UI (User Interface). When the user starts the system, the data can be received from realtime data or stored data. The system may have dynamic live data obtained instantly from the IoT devices. If not, previously-stored simulation data is loaded and relevant scenarios are run on this data.

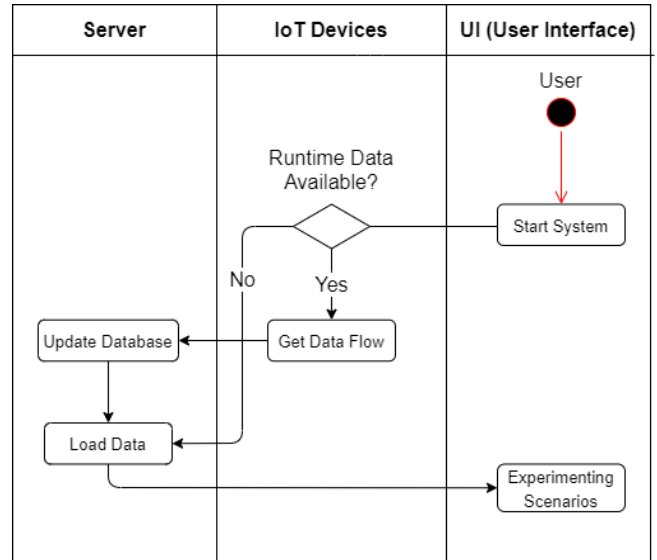

**Fig. 3:** Action Diagram of the System

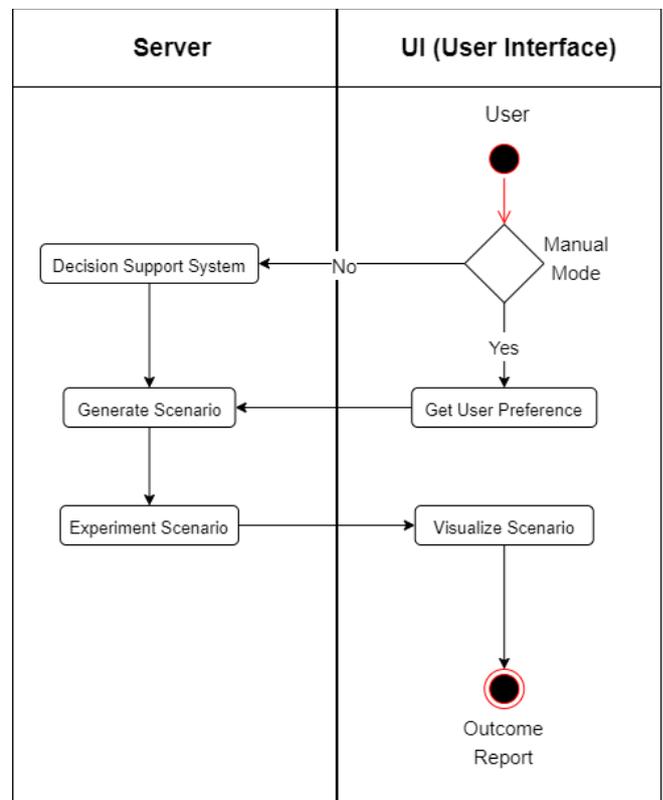

**Fig. 4:** Action Diagram of Experimenting Scenarios

Experimenting scenarios are given in detail in Fig. 4. This can be done in two ways. In manual (education) mode, choices are made by the trainee. Users' preferences are taken and according to these preferences, appropriate scenarios are created. Otherwise, the decision support system creates different scenarios by itself and decides which result is most effective in these different scenarios. Also, the decision support system can make changes to the real IoT devices. At the end of the scenario, a report is presented which includes how many people survived, state of buildings, and infrastructures. Depending on these, the effectiveness of the scenario is analyzed.

The graphical user interface is also one of the issues to be considered. It can be examined under two main options: 3D or 2D. All of these approaches have important advantages and also limitations as well. Some of the strengths and weaknesses of each system are highlighted on Table 2. 3D modelling requires too much computational power and time. Since our sources are limited, it also was not considered appropriate. Given all these situations, 2D model was chosen for the system.

**Table 2.** GUI Types

|   | Advantages | Disadvantages |
|---|---|---|
| 3D | More accurate modelling | Difficult to model very large areas |
|   | Suitable for buildings, bridges | Need more computational power |
| 2D | Large areas easily to build | Hard to model complex structures |
|   | Low computational power | Less detail, low accuracy |

## V. IMPLEMENTATION

The implementation aims to educate the emergency centre staff. The system prototype for the education mode is made where an earthquake scenario was implemented. The event of a disaster will be simulated; the situation from the beginning of the earthquake in a timeline can be analyzed by users. Post-earthquake infrastructure data can also be examined. Possible outcomes of the decisions will be reported accordingly.

A prototype simulation is implemented as a proof of concept on an ordinary laptop with 8 GB RAM, 256 GB SSD, 6 GB GTX 1050 Ti and an Intel i5-8300H 2.3 GHz CPU.

The incoming and stored data must be visualized to make it more meaningful. MapBox is selected to create the map service [18]. Conditions of the buildings and the roads are marked by visualization elements.

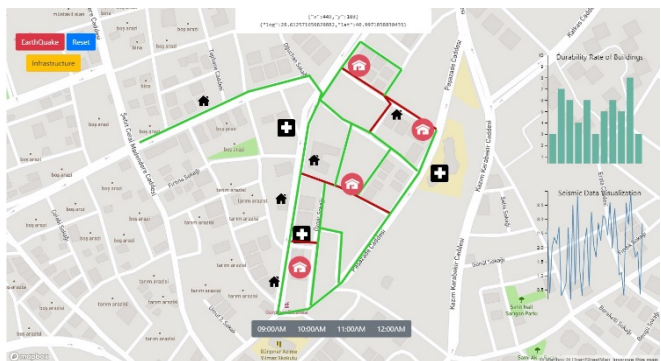

**Fig. 5:** Post-Disaster Prototype Interface

The user enters the map service and reaches the necessary live pre-disaster visualizations and analysis about buildings, rescue centres, and roads. The prototype interface is shown in Fig. 5. The user can switch to the map style containing the infrastructure information and analyzes the water, electricity, and communication infrastructures. This prototype interface is shown in Fig. 6. The environment is simulated with instant data.

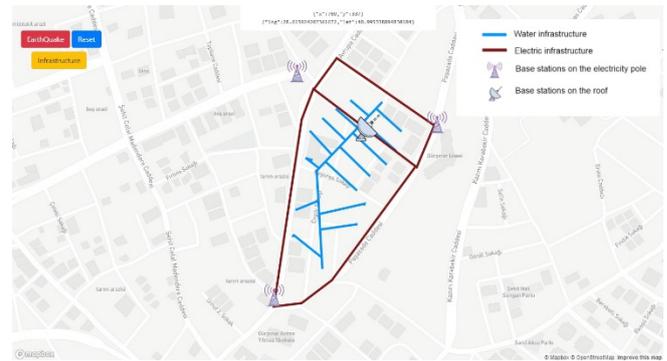

**Fig. 6:** Post-Disaster Prototype Interface for Infrastructures

The simulation environment is used to show the status of the buildings, infrastructures (such as communication, power, electricity, water and natural gas) and the roads in a timeline after the disaster. The most appropriate directions/paths for the rescue teams are given by the decision support system. The system recommends the following:

- Number and type of rescue teams according to the civilization density and the status of the buildings.

- The best live-saving scenario(s) with the most appropriate paths from the best available rescue centre to the wreck.

Simulation provides the users several options with potential success rates. When a decision is chosen, the outcome is also shown in the simulation. Then a rescue team vehicle departs from the rescue centre and goes from the shortest and safest way to save the people. The outcomes of this decision is presented to the user.

A system that uses artificial intelligence algorithms such as Breadth-First Search (BFS) [19] and Uniform-Cost Search (UCS) [20] is being developed. This will help in recommending the best options which will help rescue teams to identify better alternative to save lives.

Social media such as Twitter is an important communication channel in times of emergency. The iniquitousness of smartphones enables people to announce an emergency in real-time. Because of this feature, more agencies (i.e. disaster relief organizations and news agencies) are interested in programmatically monitoring Twitter. It's not always clear whether a person's words are announcing a disaster. An NLP (Natural Language Processing) model [21] which detects whether the tweets are fake or not was developed. Transfer learning, particularly models like Allen AI's ELMO, OpenAI's Open-GPT, and Google's BERT [22] is being used to smash multiple benchmarks with minimal task-specific fine-tuning. These are also providing the rest of the NLP community with pre-trained models that could easily (with less data and less compute time) be fine-tuned and implemented to produce state of the art results. Hence, in this project, it is decided to use Google's pre-trained BERT model.

Google's pre-trained BERT model is preferred in this project. The dataset used for training the model consists of two components. The first is the text of the tweet. The other is the estimation of whether the tweet is about a real disaster (1) or not (0). This dataset was produced using a CSV file obtained from Kaggle [23]. This file contains information; a unique identifier for each tweet, the location the tweet was sent from, and a particular keyword from the tweet. This dataset contains 7613 tweet texts. The size of the dataset is relatively small compared to modern deep learning models. Hence, the data was split into an approximately 80:10:10 ratio for the training set, development set, and test set respectively. This breaks up of 6091 training examples, 1522 validation examples, 1522 test examples. Binary cross-entropy is chosen as a loss function of this model. AdamW is chosen for optimizer of the model, because it optimizes the learning rate used in gradient descent.

## VI. RESULTS

The results of the simulation as part of our first experiments are promising. The simulation covered a small area, therefore the amount of data was limited. Building and infrastructure data was entered manually to the database. The "Earthquake Activity in Istanbul in the Last Year" dataset [24] from IBB open data was used. However, for a digital twin; the data must be provided dynamically from social media and IoT sensors. The system will be better when such an environment is provided. Using live data is to be implemented in a future study. Moreover, also the NLP model result were promising. The dataset was pretty simple and small which is not good for BERT model. It is trained just 4 epochs. After training section, accuracy on test data is about 67% that is not considered as. Increasing the size of dataset and making necessary optimizations for better performance of model is considered as a future study.

## VII. CONCLUSION

Natural and manmade disasters are an inevitable part of our daily lives and governments, institutions, local authorities and individuals must be prepared for such an event. The necessary measures before the disasters are key to reduce the damage on property and the number casualties. "Earthquakes don't kill people, buildings do" expression should not be forgotten.

Disaster management is an important issue to be addressed and digital twin technology can make it easier for us. The concept is still new for many countries, but it is expected to become mainstream within the next five to ten years. Simulations in the virtual copy can help us to learn and prepare the necessary precautions. After the disaster, even seconds are important for saving lives. It is pretty crucial to intervene immediately. Digital twins can also help us with this issue.

Test results are promising and simulation scope is planned to be extended. The simulation results are to be used to have the idea of taking necessary precautions on housing, transportation, telecommunication, natural gas, and electricity networks. The real-time data after a disaster is planned to be used on deciding how rescue operations to be carried out, figuring out the fullness/sturdiness ratio of the mission-critical buildings like hospitals.

Future works will include running this whole system in the cloud, as the amount of data from sensors and social media will increase gradually. These sensors will be placed on the buildings and infrastructures like electricity, transportation, telecommunication, and natural gas. This study will be extended to the city level and a central intelligence in which all regions facing the risk of earthquakes will be considered.